\documentclass[sigconf]{acmart}

\usepackage{graphicx}
\usepackage{amsmath}
\usepackage{graphicx}
\usepackage{capt-of}
\usepackage{xcolor}
\usepackage{amsmath,amsfonts,dsfont,pifont,bm,bbm,mathrsfs,mathtools,nicefrac}
\usepackage{algorithm,algpseudocode,listings}
\usepackage{float}
\usepackage{booktabs,multirow,adjustbox,diagbox,threeparttable}
\usepackage{colortbl}  
\usepackage{array}   
\usepackage{subfigure}
\usepackage{{hyperref}}
\usepackage[hyphenbreaks]{breakurl}
\usepackage[capitalize]{cleveref}  
\usepackage{balance} 
\crefname{section}{Sec.}{Secs.}
\Crefname{section}{Section}{Sections}
\crefname{table}{Tab.}{Tabs.}
\Crefname{table}{Table}{Tables}
\crefname{figure}{Fig.}{Figs.}
\Crefname{figure}{Figure}{Figures}
\crefname{equation}{Eq.}{Eqs.}
\Crefname{equation}{Equation}{Equations}
\newcommand{\tocite}[1]{\textcolor{red}{[TO CITE]}}

\AtBeginDocument{%
  \providecommand\BibTeX{{%
    \normalfont B\kern-0.5em{\scshape i\kern-0.25em b}\kern-0.8em\TeX}}}

\copyrightyear{2023}
\acmYear{2023}
\setcopyright{acmlicensed}
\acmConference[MM '23] {Proceedings of the 31st ACM International Conference on Multimedia}{October 29--November 3, 2023}{Ottawa, ON, Canada.}
\acmBooktitle{Proceedings of the 31st ACM International Conference on Multimedia (MM '23), October 29--November 3, 2023, Ottawa, ON, Canada}
\acmPrice{15.00}
\acmISBN{979-8-4007-0108-5/23/10}
\acmDOI{10.1145/3581783.3611763}

\begin{document}


\title{Improving Few-shot Image Generation by Structural Discrimination and Textural Modulation}

\acmSubmissionID{316}


\author{Mengping Yang}
\affiliation{%
  \institution{Key Laboratory of Smart Manufacturing in Energy Chemical Process,
  Department of Computer Science and Engineering, East China University of Science and Technology
  }
  \country{Shanghai, China}
}
\email{mengpingyang@mail.ecust.edu.cn}
\orcid{0000-0003-1503-9621}

\author{Zhe Wang}
\authornote{Corresponding author.}
\affiliation{%
  \institution{Key Laboratory of Smart Manufacturing in Energy Chemical Process,
  Department of Computer Science and Engineering, East China University of Science and Technology
  }
  \country{Shanghai, China}
}
\email{wangzhe@ecust.edu.cn}
\orcid{0000-0002-3759-2041}

\author{Wenyi Feng}
\affiliation{%
  \institution{Key Laboratory of Smart Manufacturing in Energy Chemical Process,
  Department of Computer Science and Engineering, East China University of Science and Technology
  }
  \country{Shanghai, China}
}
\email{Y10200096@mail.ecust.edu.cn}
\orcid{0000-0002-4663-0097}

\author{Qian Zhang}
\affiliation{%
  \institution{
  Department of Computer Science and Engineering, East China University of Science and Technology
  }
  \country{Shanghai, China} 
}
\email{qianzhang@ecust.edu.cn}
\orcid{0000-0003-2037-7924}

\author{Ting Xiao}
\affiliation{%
  \institution{
  Department of Computer Science and Engineering, East China University of Science and Technology
  }
  \country{Shanghai, China}
}
\email{xiaoting@ecust.edu.cn}
\orcid{0000-0003-3155-7664}

\renewcommand{\shortauthors}{Mengping Yang, Zhe Wang, Wenyi Feng, \& Qian Zhang, Ting Xiao}



\begin{abstract}








Few-shot image generation, which aims to produce plausible and diverse images for one category given a few images from this category, has drawn extensive attention.
Existing approaches either globally interpolate different images or fuse local representations with pre-defined coefficients.
However, such an intuitive combination of images/features only exploits the most relevant information for generation, leading to poor diversity and coarse-grained semantic fusion.
To remedy this, this paper proposes a novel textural modulation (TexMod) mechanism to inject external semantic signals into internal local representations.
Parameterized by the feedback from the discriminator, our TexMod enables more fined-grained semantic injection while maintaining the synthesis fidelity.
Moreover, a global structural discriminator (StructD) is developed to explicitly guide the model to generate images with reasonable layout and outline.
Furthermore, the frequency awareness of the model is reinforced by encouraging the model to distinguish frequency signals.
Together with these techniques, we build a novel and effective model for few-shot image generation.
The effectiveness of our model is identified by extensive experiments on three popular datasets and various settings.
Besides achieving state-of-the-art synthesis performance on these datasets, our proposed techniques could be seamlessly integrated into existing models for a further performance boost.
Our code and models are available at \href{https://github.com/kobeshegu/SDTM-GAN-ACMMM-2023}{here}.

\end{abstract}


\begin{teaserfigure}
\centering
  \includegraphics[width=\textwidth]{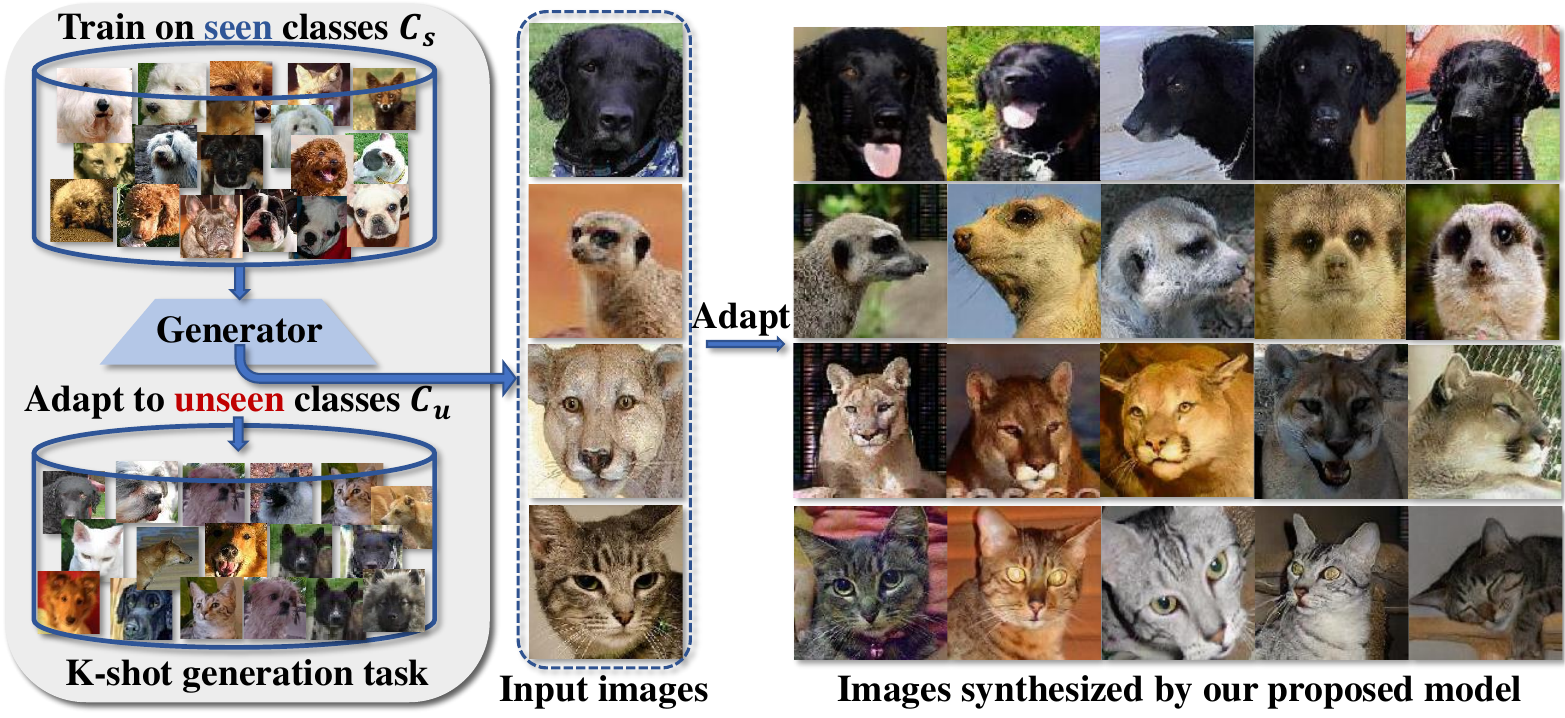}
  \vspace{-6mm}
  \caption{
  (\emph{left.}) Trained on \emph{seen} classes, the learned generator of the few-shot image generation model is then adapted to \emph{unseen} classes for producing novel images of one category given a few images (\emph{e.g.,} 1 or 3) from this category.
  (\emph{right.}) Given only a single image from one specific category, our model is capable of generating photorealistic and diverse samples.
  }
  \label{fig:teaser}
\end{teaserfigure}

\maketitle

\section{Introduction}

Thrilling features of Generative Adversarial Networks~\cite{goodfellow2014generative} such as impressive sample quality and flexible content controllability have significantly advanced visual applications including image~\cite{bigGAN, stylegan2, stylegan3, styleganxl} and video generation~\cite{styleganv, stylesv, video}, image editing~\cite{shen2020interfacegan, CeFa2021}, image-to-image translation~\cite{richardson2021encoding, yang2022unsupervised, liu2019few, saito2020coco}, \emph{etc.}
However, their breakthroughs mainly attribute to ample training data and sufficient computation resources.
For instance, current state-of-the-art StyleGAN models~\cite{stylegan, stylegan2, stylegan3} are trained on Flickr-Faces (FFHQ) which involves 70$K$ images for desirable performance.
Such requirement on massive data poses limitations of GANs on adapting to new categories~\cite{Bartunov2018FewshotGM, liang2020dawson} and practical domains with only limited training data~\cite{hong2020f2gan, stylegan2-ada, InsGen}.
%
Consequently, it is critical to consider how to produce novel images given only a few images per category.
Such a task, dubbed few-shot image generation~\cite{hong2020deltagan, wavegan, ding2022attribute, ding2023stable}, has attracted extensive attentions recently.
%

The goal of few-shot image generation is to quickly adapt knowledge learned from seen classes to unseen classes (see~\cref{fig:teaser}).
Specifically, the model is firstly trained in an episodic manner~\cite{vinyals2016matching} on seen categories with sufficient training samples and per-sample class labels.
Then, the learned model is required to transfer the generation ability to a new unseen category, \emph{i.e.,} producing diverse images for a new class given a few images (\emph{e.g.,} 3) from the same class, and there are no overlaps between the seen categories and the unseen categories.
Thus the model is expected to learn how to generate novel images instead of merely capturing the distribution of seen classes.

Existing few-shot generation models seek to ameliorate the synthesis quality via 1) transforming intra-class representations to new classes~\cite{antoniou2017data}, 2) optimizing new criterion to achieve better knowledge transferabilities~\cite{clouatre2019figr, liang2020dawson, bartunov2018few}, and 3) fusing global images or local features~\cite{hong2020f2gan, lofgan, wavegan}.
For instance, LoFGAN~\cite{lofgan} produces plausible and diverse images by fusing the local features of different images based on a pre-defined similarity map.
The current state-of-the-art WaveGAN~\cite{wavegan} encourages the model to synthesize high-frequency signals with frequency residual connections, enabling better awareness of spectrum information.
Although these models have made remarkable progress, they still struggle to produce images with
desirable diversity and fidelity simultaneously due to two critical limitations.
On one hand, they only fuse semantically relevant features \emph{i.e.}, features with relatively high similarity, lacking more fine-grained semantic combinations and thus losing diversity.
On the other hand, the arrangement of generated content might be arbitrary after fusing the local features since no explicit structural guidance is provided, degrading the synthesis fidelity.

%
we present a novel few-shot generation model, dubbed SDTM-GAN, that addresses the aforementioned
limitations through the incorporation of two key components: structural discrimination (StructD) and textural
modulation (TexMod). 
%
%
Specifically, TexMod is performed via modulating the textural style of generated images at the semantic level.
By injecting external semantic layouts from different samples into the internal textural style of generated images, TexMod could better combine local semantic representations and thus capture more semantic variations.
Considering that fusing semantic features might cause arbitrary structures, we furthre develop StructD to ensure global coherence.
Concretely, we first perform Laplace Operator~\cite{van1989nonlinear} on the input images to obtain laplacian representations which encode rich global structural information such as contour edges and object boundaries.
A lightweight discriminator, \emph{i.e.,} StructD, which distinguishes the laplacian representations of real and generated images, is then proposed to explicitly provide structural guidelines to the generator, facilitating the fidelity of global appearance.
Meanwhile, inspired by the findings that neural networks prefer to fit low-frequency signals while tending to ignore high-frequency information~\cite{xu2019frequency, gao2021high, wang2022fregan}, we further adopt a frequency discriminator to encourage the discriminator to capture high-frequency signals.
%

Together with the above techniques,  our model can 1) capture the global structural and high-frequency signals, facilitating the fidelity of generated images;
and 2) produce diverse images via modulating semantic features in a more fine-grained manner.
We evaluate the effectiveness of our method on several popular few-shot datasets and the results demonstrate that our method achieves appealing synthesis performance in terms of image quality and richness (see~\cref{fig:teaser} and~\cref{sec:experiments}).
Additionally, our proposed techniques are complementary to existing models, \emph{i.e.,} integrating our methods into existing models gains a further performance boost.


\noindent \textbf{Contributions.}
Our contributions are summarized as follows:
1) We propose a novel few-shot image generation model (\emph{i.e.,} SDTM-GAN) which incorporates structural discrimination and textural modulation to respectively improve the global coherence of generated images and accomplish more fine-grained semantic fusion.
2) The proposed techniques could be readily integrated into existing few-shot generation models to further boost the performance with negligible computation cost, further suggesting the efficacy and compatibility of our methods. 
3) Under popular benchmarks and various experimental settings, our method consistently outperforms prior arts by a substantial margin. 
Besides, the images produced by our model are utilized for augmenting the training set for downstream classification problems, leading to improved classification accuracy.
Overall, our method brings advantageous potential for improving few-shot image generation and downstream applications.
\vspace{-2mm}
\section{Related Works}

\noindent \textbf{Generative adversarial networks} (GANs)~\cite{goodfellow2014generative, yangimproving} are typically composed of a discriminator and a generator, where the former learns to distinguish real images from generated ones and the latter tries to deceive the discriminator via reproducing the data distribution.
Benefiting from the compelling ability to capture data distributions, GANs have been ubiquitously applied in various visual domains, such as image-to-image translation~\cite{yang2022unsupervised, liu2019few}, image/video generation~\cite{stylesv, styleganv, bigGAN}, image manipulation and inpainting~\cite{CeFa2021, dong2022incremental, wang2022diverse}, \emph{etc.}
However, their performance drops drastically when trained on few-shot datasets due to the discriminator overfitting and memorization issues~\cite{stylegan2-ada, DiffAug, li2022fakeclr, li2022comprehensive}.
Some recent works mitigate the overfitting problem by applying extensive data augmentation~\cite{jiang2021deceive, stylegan2-ada, DiffAug} to enlarge the training sets or developing additional branches and constraints~\cite{liu2021towards, InsGen, wang2020minegan, wang2022fregan, yang2023protogan, saxena2023re, li2023systematic} to dig more available information.
Unlike their concentration on improving unconditional image generation, our goal is to produce novel images for one specific class when provided with a few images from the same class.
Trained in an episodic manner as few-shot learning~\cite{vinyals2016matching, hong2020f2gan, ding2022attribute}, our model is expected to capture the knowledge of generating new images.

\noindent \textbf{Few-shot image generation.}
Many attempts have been endowed to ameliorate synthesis quality for few-shot scenarios.
Existing alternatives could be roughly divided into three categories based on their different techniques~\cite{hong2020f2gan, hong2020deltagan, ding2022attribute, ding2023stable, hong2022few}, namely optimization-based, fusion-based, and transformation-based approaches.
Optimization-based methods~\cite{clouatre2019figr, liang2020dawson} combine GANs with meta-learning~\cite{finn2017model} to generate new images via finetuning the parameters of the inner generating loop and outer meta training loop, but their sample quality is often limited.
Differently, transformation-based models like DAGAN~\cite{antoniou2017data} transform intra-class and randomly sampled latent representations into new images, enabling relatively high diversity yet bringing unsatisfactory aliasing artifacts.
By contrast, the fusion-based~\cite{hong2020f2gan, lofgan} methods achieve better synthesis quality.
For instance, F2GAN~\cite{hong2020f2gan} proposes a fusing-and-filling scheme to interpolate input conditional images and fill fine details into the fused image.
Considering that fusing the image globally leads to a semantic misalignment, LoFGAN~\cite{lofgan} further improves the performance by combining local representations following a pre-computed semantic similarity.
Moreover, WaveGAN~\cite{wavegan} explicitly encourages the model to pour more attention on high-frequency signals, which previous models usually ignore.

However, there are still two main limitations that remain underexplored among prior studies.
On one hand, fusing local features based on a similarity map only combines the most relevant semantics, leading to unfavorable synthesis diversity.
Besides, no learnable parameters are involved in the fusion process, lacking explicit optimization.
%
On the other hand, global coherence might be affected by local fusion and produce arbitrary images without global structural guidelines.
In this paper, we fuse local semantics via learnable textural modulation and explicitly provide structural information to the model.

\noindent \textbf{Frequency bias in GANs.}
Deep neural networks are identified to have a preference for capturing frequency 
signals from low to high~\cite{xu2019frequency, schwarz2021frequency}, which also holds for GANs.
Accordingly, many works have been developed to improve GANs` frequency awareness.
For instance, Jiang \emph{et al.} propose focal frequency loss to iteratively attach higher importance to hard frequency signals~\cite{jiang2021focal}.
Gao \emph{et al.} alleviate GAN's frequency bias by residual frequency connections~\cite{gao2021high} and Yang \emph{et al.} employ high-frequency discriminator~\cite{wang2022fregan} to achieve this.
Similarly, we assign a frequency discriminator to help the model better encode frequency signals.

\noindent \textbf{Modulation techniques} are effective ways to combine external information with internal representations and have been successfully applied to many practical domains such as style transfer~\cite{huang2017arbitrary}, semantic image synthesis and editing~\cite{luo2022context, park2019semantic, perez2018film}.
Specifically, the input features are first normalized to zero mean and unit deviation.
Then, the normalized representations are modulated by injecting external signals from other features.
In this way, the modulated features contain original content while capturing external semantic layouts.
Following this philosophy, we apply this to few-shot image generation and develop a two-branch textural modulation to fuse local features in a more fine-grained manner.
By incorporating internal textural content with external semantic representations through learnable modulating parameters, our model promotes a more diverse generation.
Details will be given in the next section.

\begin{figure*}
    \centering
    \includegraphics[width=\linewidth]{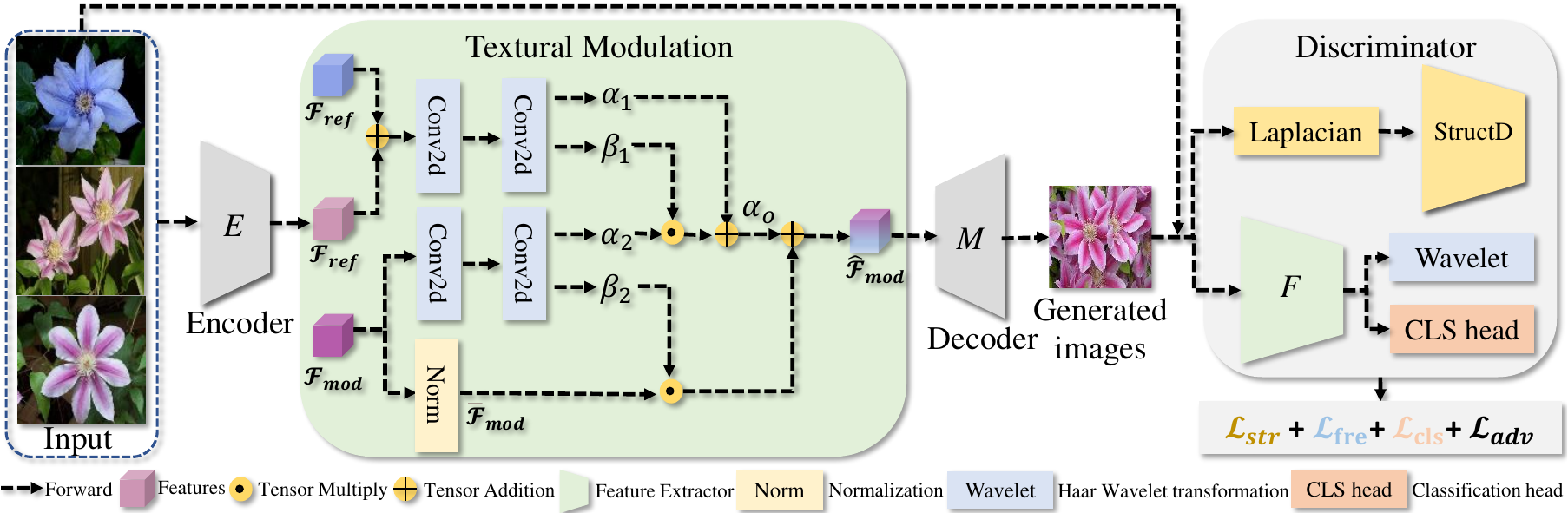}
    \vspace{-10pt}
    \caption{The overall pipeline of our model. Textural modulation (TexMod) enables more fine-grained semantic fusion via injecting the outer semantic information into the inner representations.
    Structural discriminator (StructD) explicitly encourages the model to capture the global structural signals, ensuring more reliable and reasonable synthesis.
    } 
    \label{fig:framework}
\end{figure*}
\vspace{-2mm}

\section{Methodology}
%
%
In this section, we present the technical detail the proposed methods, namely structural discriminator (StructD) and textural modulation (TexMod).
The formulation of few-shot image generation and our overall framework are presented in~\cref{sec:problem_definition}, followed by the description of our StructD and TexMod respectively in~\cref{sec:structD} and~\cref{sec:TexMod}.
Finally, \cref{sec:optimization} presents the optimization objectives.

\subsection{Preliminary and Overview}
\label{sec:problem_definition}

\noindent \textbf{Preliminary.}
~\cref{fig:teaser} shows the setting of few-shot image generation.
Concretely, the model is first trained on seen classes $\mathcal{C}_s$ in an episodic manner.
Episodic training is achieved by feeding $N$-way-$K$-shot images as input for each iteration, where $N$ denotes the number of classes and $K$ is the number of images for each class.
Such a paradigm makes the model capture transferable ability for image generation.
Then, the model is expected to produce novel images given several images from unseen classes $\mathcal{C}_u$ ($\mathcal{C}_u \cap \mathcal{C}_s = \emptyset$).

\noindent \textbf{Overall framework.}
~\cref{fig:framework} illustrates the overall framework of our proposed model.
The generator consists of one encoder ($E$) and one decoder ($M$), the former projects input images to latent features and the latter decodes the modulated representations to produce new images.
Textural modulation (TexMod) enables more detailed fusion by injecting the outer semantic layout into inner textures with learnable parameters.
Besides, by leveraging the Laplacian representations as a global guidance, the model can eliminate productions with discordant structures.
%

\subsection{Textural Modulation}
\label{sec:TexMod}
%
Textural modulation (TexMod) injects external semantic information into internal features.
~\cref{fig:framework} shows the pipeline of TexMod given three input images from each category.
Firstly, $K$ features $\mathbf{F} = \mathcal{F}_k|_{k=1}^K$, $\mathcal{F}_k \in \mathcal{R}^{w\times h \times c}$($K = 3$ here), where $w, h, c$ denote the feature dimensions, are obtained from the encoder $E$.
Then, one feature $\mathcal{F}_{mod}$ for modulation is randomly chosen and the other referenced features $\mathcal{F}_{ref}$ are used for injection.
Finally, the modulated feature is obtained following a two-stage injection mechanism.

\noindent \textbf{First-stage injection.}
In order to obtain reasonable modulation weights for semantic injection, we perform 2d convolutions on the chosen feature $\mathcal{F}_{mod}$ and the sum of reference features $\mathcal{F}_{ref}$ respectively, obtaining two sets of modulation parameters ($\alpha_1$, $\beta_1$) and ($\alpha_2$, $\beta_2$).
The 2d convolution here encodes semantic information of local features and generates learnable parameters, enabling more controllable and fine-grained fusion.
The first stage semantic of injection is then accomplished by
\begin{equation}
    \alpha_{o} = (\mathbf{1} + \beta_1) \bigodot \alpha_2 + \alpha_1,
\end{equation}
where $\bigodot$ demotes the element-wise multiplication and $\alpha_o$ is the obtained parameter for the second-stage modulation.
All parameters share the same dimension with the chosen feature $\mathcal{F}_{mod}$.

\noindent \textbf{Second-stage injection.}
Stage one injects the semantic representations of referenced features into that of the chosen feature $\mathcal{F}_{mod}$. 
However, the overall texture might be overridden by semantic fusion.
Accordingly, we first obtain the normalized feature ($\bar{\mathcal{F}}_{mod}$) by normalizing the chosen feature.
Then, the modulated parameter $\alpha_{o}$ and $\beta_2$ are leveraged for a second-stage injection on $\bar{\mathcal{F}}_{mod}$:
\begin{equation}
    \hat{\mathcal{F}}_{mod} = (\mathbf{1} + \beta_2) \bigodot \bar{\mathcal{F}}_{mod} + \alpha_o,
\end{equation}
where $\hat{\mathcal{F}}_{mod}$ is the output feature which maintains the texture of $\mathcal{F}_{mod}$ meanwhile encodes rich semantic details of referenced features $\mathcal{F}_{ref}$.
Additionally, the feature for modulation is randomly chosen at each training episodic, involving more semantic variance for injection.
Finally, the modulated feature $\hat{\mathcal{F}}_{mod}$ is forwarded into the decoder $M$ to synthesize new images.
Through the proposed two-stage modulation, more fine-grained semantic injection is achieved since all semantic information of referenced features is integrated into semantic fusion, improving the diversity.
Moreover, the modulation weights are optimized following the feedback of the discriminator, ensuring fidelity is not compromised.
%

\subsection{Structural and Frequency Discriminator}
\label{sec:structD}
Typically, existing approaches perform adversarial loss and classification loss to penalize the discriminator.
However, the overall structure and outline of generated images might be arbitrary without explicit global guidance.
We ameliorate this by enforcing the discriminator to capture global structural information.
Specifically, the Laplacian operation is first leveraged to extract the global structural signals (\emph{e.g.,} contour edges and object boundaries).
Laplacian operation is accomplished via a convolutional layer with the Laplacian kernel:
%
%
%
\begin{equation}
\mathrm{Kernel}_{Laplacian}=\left[\begin{array}{ccc}
0 & -1 & 0 \\
-1 & 4 & -1 \\
0 & -1 & 0
\end{array}\right].
\end{equation}
The Laplacian kernel is utilized to project input images to Laplacian representations, then a structural discriminator (StructD) is employed to encode global signals.
The losses of StructD are defined as:
\begin{equation}
\begin{aligned}
    \mathcal{L}_{str}^{D} &= \max (0,1-D_{str}(\mathbf{x}))+\max (0,1+D_{str}(\hat{\mathbf{x}})), \\
    \mathcal{L}_{str}^{G} &= -D_{str}(\hat{\mathbf{x}}),
\end{aligned}
\end{equation}
where $D_{str}$ represents StructD. $\mathbf{x}$ and $\hat{\mathbf{x}}$ is input real and generated images, respectively.
Akin to conventional discriminators, StrctD is comprised of convolutional and activation layers.
Notably, only encoding structural signals, our StrctD is lightweight and introduces negligible(see~\cref{tab:costs}) additional computation burdens.

\noindent \textbf{Frequency discriminator.}
In order to mitigate the model's frequency bias, we employ wavelet transformation on the extracted features and obtain high-frequency signals of input images.
Then we encourage the model to distinguish high-frequency signals of real images from that of generated samples, forming a frequency discriminator which improves the frequency awareness of our model.
The frequency losses are given by
\begin{equation}
\begin{aligned}
    \mathcal{L}_{fre}^{D} &= \max (0,1-D_{fre}(\mathcal{H}(F(\mathbf{x}))))+\max (0,1+D_{fre}(\mathcal{H}(F(\hat{\mathbf{x}})))), \\
    \mathcal{L}_{fre}^{G} &= -D_{fre}(\mathcal{H}(F(\hat{\mathbf{x}}))),
\end{aligned}
\end{equation}
where $F$ is the feature extractor, and $\mathcal{H}$ represents the Haar wavelet transformation~\cite{daubechies1990wavelet} that decomposes features into different frequency components.
The obtained high-frequency signals are then forwarded into the frequency discriminator $D_{fre}$, which contains an adaptive-average-pool and a Conv2D layer for calculation.

\subsection{Optimization}
\label{sec:optimization}

Two subnetworks are involved for optimization in our model, namely generator ($G$) and discriminator ($D$), and $G$ and $D$ are optimized alternatively in an adversarial manner.
Formally, let $\mathbf{X} = \{\mathbf{x_1}, \mathbf{x_2}, \mathbf{x_3}, ...\}$ demotes the input real images and $\mathbf{c}(\mathbf{x_i})$ is the corresponding labels for $\mathbf{x_i}$ (only for $\mathcal{C}_s$).
Image produced by G is denoted as ${\mathbf{\hat{x}} = G (\mathbf{X})}$, which $D$ seeks to distinguish from real images by computing $D (\mathbf{X})$.

\noindent \textbf{Adversarial loss.}
The hinge version of adversarial loss is employed for training.
$D$ tries to assign higher scores for real images while lower ones for generated samples, and $G$ seeks to produce plausible images to fool $D$:
\begin{equation}
\begin{aligned}
\mathcal{L}_{{adv}}^D & =\max (0,1-D(X))+\max (0,1+D(\hat{x})), \\
\mathcal{L}_{{adv}}^G & =-D(\hat{x}).
\end{aligned}
\end{equation}

\noindent \textbf{Classification loss} ensures the model to capture the class distribution of training sets (\emph{i.e.,} seen classes $\mathcal{C}_s$).
Such that, the model could produce images for one category given the labeled class.
Formally, classification loss is calculated by
\begin{equation}
\begin{aligned}
\mathcal{L}_{c l s}^D & =-\log P(c(\mathbf{x}) \mid \mathbf{x}), \\
\mathcal{L}_{c l s}^G & =-\log P(c(\hat{\mathbf{x}}) \mid \hat{\mathbf{x}}),
\end{aligned}
\end{equation}
where $P(\cdot)$ denotes the sample's probability of belonging to class $c$.


Consequently, the generator $G$ and the discriminator $D$ are respectively trained by combining the above losses linearly.
\begin{equation}
\begin{aligned}
\mathcal{L}_D & =\mathcal{L}_{{adv}}^D+\mathcal{L}_{c l s}^D+\lambda_{fre}\mathcal{L}_{fre}^{D}+\lambda_{str}\mathcal{L}_{str}^{D}, \\
\mathcal{L}_G & =\mathcal{L}_{{adv}}^G+\mathcal{L}_{c l s}^G+\lambda_{fre}\mathcal{L}_{fre}^{G}+\lambda_{str}\mathcal{L}_{str}^{G}.\\
\end{aligned}
\end{equation}
Note that in our implementation, $\lambda_{fre} = \lambda_{str} = 1$, and the detailed comparisons are presented in~\cref{sec:ablation_studies}.
\section{Experiments}
\label{sec:experiments}

\subsection{Experimental Setup}

\noindent \textbf{Datasets.}
We evaluate the effectiveness of the proposed method on three popular datasets, namely Flowers~\cite{flowers}, Animal Faces~\cite{liu2019few}, and VGGFaces~\cite{cao2018vggface2}.
These datasets are devided into seen ($\mathcal{C}_s$) and unseen ($\mathcal{C}_u$) classes respectively for training and testing as in~\cite{hong2020f2gan, wavegan, lofgan}.
~\cref{tab:dataset} provides the detailed splits of these datasets.

\noindent \textbf{Evaluation metrics and baselines.}
Fréchet Inception Distance (FID)~\cite{FID} and Learned Perceptual Image Patch Similarity (LPIPS)~\cite{zhang2018unreasonable} serve as the quantitative metric for comparison.
%
FID reflects the synthesis quality via computing the similarity between the generated distribution and the real distribution, and lower FID indicates better performance.
LPIPS delivers sample diversity by capturing the variation of generated images, and higher LPIPS means better diversity.
%
Moreover, we leverage LoFGAN~\cite{lofgan} and WaveGAN~\cite{wavegan} as baselines and implement our proposed techniques upon their official code for evaluation.
%
Noticeably, all evaluations strictly follow the prior arts~\cite{hong2020f2gan, wavegan, lofgan} for a fair comparison.


\begin{table}
\caption{The splits of seen/unseen images (``img'') and classes (``cls'') on three datasets.} 
\centering
\vspace{-6pt}
\resizebox{0.7\columnwidth}{!} {
\begin{tabular}{l|lr|lr}  
\toprule[0.8pt]
\multirow{2}{*}{Dataset} & \multicolumn{2}{c|}{Seen} &\multicolumn{2}{c}{Unseen}\cr
                         &\#cls  & \#img    &\#cls  & \#img \cr \hline
            Flowers      & 85    & 3400     & 17    & 680   \cr
            Animal Faces & 119   & 11900    & 30    & 3000  \cr
            VGGFace      & 1802  & 180200   & 552   & 55200 \cr
\bottomrule[0.8pt]
\end{tabular}
}
\label{tab:dataset}
\end{table}

\noindent \textbf{Implementation Details.}
%
TexMod is implemented with four convolutional layers as shown in~\cref{fig:framework} to obtain modulated parameters. 
The input and output of each convolutional layer have the same dimension, facilitating the injection of semantic features.
As for the StrctD, two convolutional layers and one adaptive-average-pooling layer are employed to encourage the model to capture the global layout and outline of images.
The model is trained for $100K$ iterations and the last checkpoint is used for evaluation.
For each iteration, $K$ (\emph{e.g.,} 1, 3) conditional images from one category randomly sampled from seen classes $\mathcal{C}_s$ are used for training.
Adam optimizer~\cite{kingma2014adam} is used and the batchsize is $8$.
The learning rates for $G$ and $D$ are set to $0.0001$ for the first half iterations, and decay to $0$ linearly for the next $50K$ iterations.
All experiments are conducted on one NVIDIA 3090 with 24G memory and implemented with the PyTorch framework.

\begin{figure*}
    \centering
    \vspace{-5pt}
    \includegraphics[width=.8\textwidth]{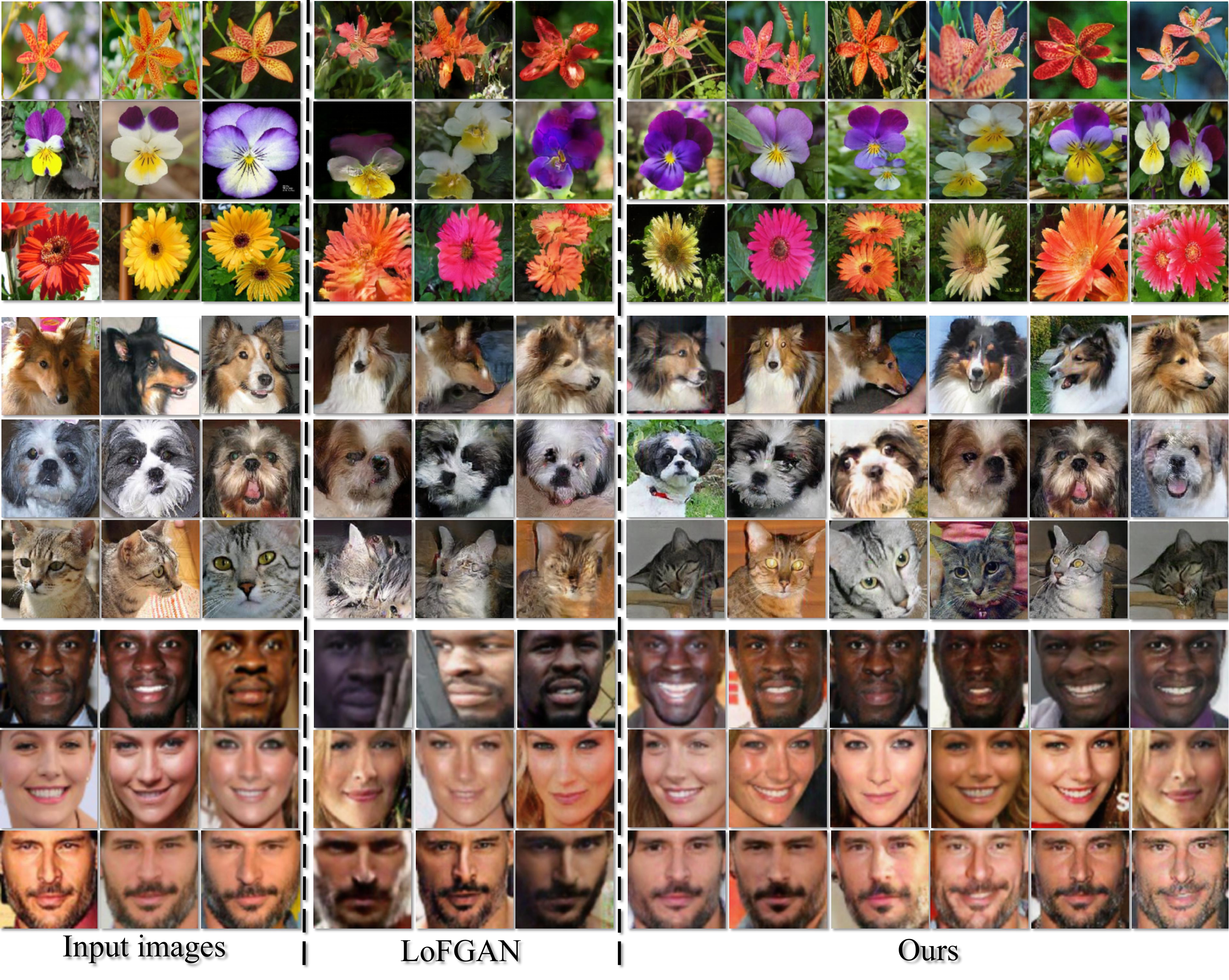}
    \vspace{-5pt}
    \caption{Qualitative comparison results of our method with LoFGAN. Images produced by our model performs better in term of the global structure (\emph{e.g.,} the outline and shape of petals and the coherence of Animal Faces) and semantic variance (\emph{e.g.,} different hair colors of Animal Faces and various expression of Human Faces).}
    \label{fig:qualitative}
    \vspace{-5pt}
\end{figure*}

\setlength{\tabcolsep}{4pt}
\begin{table*}
\vspace{-6pt}
\caption{Comparisons of FID ($\downarrow$) and LPIPS ($\uparrow$) scores on images generated by different methods for unseen categories. The marked results marked with different colors denote we evaluate our methods based on the top of their official implementations.} 
\centering
\resizebox{1.4\columnwidth}{!} {
\begin{tabular}{l|l|rr|rr|rr}
\toprule
\multirow{2}{*}{Method} & \multirow{2}{*}{Setting} & \multicolumn{2}{c|}{Flowers} & \multicolumn{2}{c|}{Animal Faces} & \multicolumn{2}{c}{VGGFace}  \cr
& & FID ($\downarrow$)  & LPIPS ($\uparrow$) &FID ($\downarrow$) & LPIPS ($\uparrow$) & FID ($\downarrow$)  & LPIPS ($\uparrow$)  \cr 
\cmidrule(r){1-1} \cmidrule(r){2-2} \cmidrule(r){3-4}  \cmidrule(r){5-6}  \cmidrule(r){7-8} 
FIGR~\cite{clouatre2019figr}    & 3-shot & 190.12 & 0.0634 & 211.54  & 0.0756 & 139.83 & 0.0834 \cr
GMN~\cite{bartunov2018few}      & 3-shot & 200.11 & 0.0743 & 220.45  & 0.0868 & 136.21 & 0.0902 \cr
DAWSON~\cite{liang2020dawson}   & 3-shot & 188.96 & 0.0583 & 208.68  & 0.0642 & 137.82 & 0.0769 \cr
DAGAN~\cite{antoniou2017data}   & 3-shot & 151.21 & 0.0812 & 155.29  & 0.0892 & 128.34 & 0.0913 \cr
MatchingGAN~\cite{matchinggan}  & 3-shot & 143.35 & 0.1627 & 148.52  & 0.1514 & 118.62 & 0.1695 \cr
F2GAN~\cite{hong2020f2gan}      & 3-shot & 120.48 & 0.2172 & 117.74  & 0.1831 & 109.16 & 0.2125 \cr
DeltaGAN~\cite{hong2020deltagan}& 3-shot & 104.62 & 0.4281 & 87.04   & 0.4642 & 78.35  & 0.3487 \cr
FUNIT~\cite{liu2019few}         & 3-shot & 100.92 & 0.4717 & 86.54   & 0.4748 & -      & -      \cr
DiscoFUNIT~\cite{hong2022few}              & 3-shot & 84.15  & \textbf{0.5143} & 66.05   & 0.5008 & -      & -      \cr
SAGE~\cite{ding2023stable}                    & 3-shot & 41.35  & 0.4330 & 27.56   & \textbf{0.5451} & 32.89  & 0.3314 \cr \midrule
\rowcolor{blue!10} LoFGAN~\cite{lofgan} & 3-shot & 79.33 & 0.3862 & 112.81   & 0.4964 & 20.31  & 0.2869 \cr
\rowcolor{blue!10} $+$ Ours     & 3-shot & \textbf{74.08} & \textbf{0.3983} & \textbf{96.74}   & \textbf{0.5028} & \textbf{12.28}  & \textbf{0.3203} \cr 
\midrule
\rowcolor{green!10} WaveGAN~\cite{wavegan} & 3-shot & 42.17 & 0.3868 & 30.35   & 0.5076 & 4.96  & 0.3255 \cr
\rowcolor{green!10} $+$ Ours     & 3-shot & \textbf{39.51} & \textbf{0.3970} & \textbf{26.65}   & \textbf{0.5109} & \textbf{3.96}  & \textbf{0.3346} \cr 
\midrule
DAGAN~\cite{antoniou2017data}    & 1-shot & 179.59 & 0.0496   & 185.54 & 0.0687  & 134.28  & 0.0608 \cr
DeltaGAN~\cite{hong2020deltagan} & 1-shot & 109.78 & 0.3912   & 89.81  & 0.4418  & 80.12  & 0.3146 \cr
FUNIT~\cite{liu2019few}          & 1-shot & 105.65 & 0.4221   & 88.07  & 0.4362  & -  & - \cr
DiscoFUNIT~\cite{saito2020coco}  & 1-shot & 90.12  & \textbf{0.4436}   & 71.44  & 0.4411  & -  & - \cr \midrule
\rowcolor{blue!10} LoFGAN~\cite{lofgan} & 1-shot & 137.47 & 0.3868 & 152.99   & 0.4919 & 26.89  & 0.3208 \cr
\rowcolor{blue!10} $+$ Ours      & 1-shot & \textbf{124.74} & \textbf{0.3900} & \textbf{147.87}   & \textbf{0.4925} & \textbf{25.17}   & \textbf{0.3267} \cr 
\midrule
\rowcolor{green!10} WaveGAN~\cite{wavegan} & 1-shot & 55.28 & 0.3876 & 53.95   & 0.4948 & 12.28  & 0.3203 \cr
\rowcolor{green!10} $+$ Ours        & 1-shot & \textbf{52.89} & \textbf{0.3924} & \textbf{50.04}   & \textbf{0.5002} & \textbf{9.27}  & \textbf{0.3214} \cr
\bottomrule
\end{tabular}
}
\label{tab:performance_metric}
\end{table*}

\subsection{Quantitative Results}

\noindent \textbf{Three-shot image generation.}
The upper part of \cref{tab:performance_metric} presents the comparison on 3-shot image generation tasks.
Obviously, our proposed techniques bring consistent performance boosts under all tested datasets and baselines.
For instance, our proposed techniques improve the FID and LPIPS scores of LoFGAN (\emph{resp.}, WaveGAN) on VGGFace from $20.31$ (\emph{resp.}, $4.96$) to $12.28$ (\emph{resp.}, $3.96$) and from $0.2869$ (\emph{resp.}, $0.3255$) to $0.3203$ (\emph{resp.}, $0.3346$).
Despite being evaluated on different baselines, \emph{i.e.,} WaveGAN and LoFGAN, the proposed approach continuously improves the synthesis quality.
For instance, by integrating our proposed techniques with WaveGAN, new state-of-the-art FID scores on all tested datasets are established, \emph{i.e.,} $39.51$, $26.65$, and $3.96$ respectively on Flowers, Animal Faces, and VGGFace.
Regarding the LPIPS score, our proposed techniques also consistently gain improvements with respect to different datasets and baselines.
Such observations indicate the positive potential of our method for few-shot image generation.

\noindent \textbf{One-shot image generation.}
When it comes to one-shot image generation, the fusion strategy might not work since only one input image is employed for generating novel images.
We continue to use the implementations of LoFGAN and WaveGAN without fusion blocks for one-shot image generation tasks.
The bottom part of~\cref{tab:performance_metric} shows the quantitative results.
Still, the synthesis performance under one-shot settings is substantially improved by our proposed techniques.
Concretely, on WaveGAN, our method improves the FID from 55.28 to 52.89 ($\downarrow$ 4.3\%), 53.95 to 50.05 ($\downarrow$ 7.2\%), and 12.28 to 9.27 ($\downarrow$ 24.51\%) on Flower, Animal Faces and VGGFace respectively.
Additionally, LPIPS scores also gain effective improvements under all settings, further demonstrating the effectiveness of our method.

The effectiveness of our proposed method is identified via combining them with different baselines (\emph{i.e.,} LoFGAN and WaveGAN) for different tasks (\emph{i.e.,} three-shot and one-shot generation).
Our method consistently gains substantial boosts on synthesis fidelity and diversity under all settings.
Namely, the proposed techniques indeed improve the synthesis quality and are complementary to existing approaches, which further manifests the compatibility.

\begin{table}
\small
\caption{Computational cost comparisons. Our model introduces ignorable computation burden.}
\vspace{-6pt}
\centering
{
\begin{tabular}{c|c|c|c}
\toprule
Method & \# params &  FLOPS & training time(h)\\ 
\cmidrule(r){1-1} \cmidrule(r){2-2} \cmidrule(r){3-3} \cmidrule(r){4-4}
LoFGAN  & 39.35M    & 139.47G & 23.28\\
WaveGAN & 39.33M    & 139.24G & 23.17\\
+Ours   & 40.30M    & 143.12G & 23.63\\
\bottomrule
\end{tabular}
}
\label{tab:costs}
\end{table}

\noindent \textbf{Computational cost.}
~\cref{tab:costs} provides the computational burdens of our method with respect to the parameter amount, FLOPS, and training time.
Clearly, our method introduces negligible costs ($\uparrow$ \%2.47) compared with LoFGAN and WaveGAN, while significantly improving the synthesis quality under various settings.

\begin{table}
\centering
\small
\vspace{-2mm}
\caption{Classification accuracy of augmentation.
``Base" denotes no augmentation is performed.}
\vspace{-6pt}
\begin{tabular}{c|c|c|c|c}
\toprule
Datasets      & Base     & LoFGAN   & WaveGAN  & Ours         \\ 
\cmidrule(r){1-1} \cmidrule(r){2-2} \cmidrule(r){3-3} \cmidrule(r){4-4}  \cmidrule(r){5-5}
Flowers        & 64.71    & 80.78  & 84.71    & \textbf{86.09}           \\
Animal Faces      & 20.00    & 26.10  & 32.19    & \textbf{33.38}           \\
VGGFace       & 50.76    & 64.74  & 77.36    & \textbf{79.17}           \\  \bottomrule
\end{tabular}
\label{tab:clsresults}
\end{table}

\subsection{Qualitative results}
Here we qualitatively investigate the synthesis quality of our model.
To be specific, after trained on seen classes $\mathcal{C}_s$ in an episode way (\emph{i.e.,} providing $K$ images from each class for training), the model is expected to produce novel images for a category given a few images from this category. 
Both three-shot and one-shot generation tasks are involved for a more reliable evaluation.

~\cref{fig:teaser} and~\cref{fig:qualitative} provide the generated images of our method for one-shot and three-shot generation tasks respectively.
It can be seen that our model could generate diverse and photorealistic images, even when only one input image is available.
Besides, compared with images generated by LoFGAN, the overall outline, and structure of images synthesized by our model are more reasonable and plausible.
For instance, our model performs significantly better regarding the outline and shape of petals and the coherence of Animal Faces.
Furthermore, our model could produce images with rich semantic variances in terms of color, style, and texture, facilitating more diverse output.
Namely, with delicate designs toward the global structure and textural modulation, our model gains convincing improvements in generation quality.
More results can be found in the appendix.

\subsection{Augment for Downstream Classification}
%
We further evaluate the synthesis quality by augmenting the training sets with generated images for downstream classification problems.
Firstly, a ResNet-18 model is pre-trained on seen classes.
Then, the unseen classes are divided into $\mathcal{D}_{train}$, $\mathcal{D}_{val}$, and $\mathcal{D}_{test}$ respectively.
The pre-trained ResNet-18 is further trained on $\mathcal{D}_{train}$ (\emph{i.e.,} Base in~\cref{tab:clsresults}) and tested on $\mathcal{D}_{test}$.
Finally, we augment $\mathcal{D}_{train}$ by generating samples with our model to obtain $\mathcal{D}_{aug}$ for comparison, the augmentation amount for Flowers, Animal Faces, and VGGFace are respectively 30, 50, and 50.

~\cref{tab:clsresults} showcases the classification results.
As could be seen from the results, our model achieves higher accuracy (\emph{i.e., }$86.09$, $33.38$, and $79.17$ respectively on Flowers, Animal Faces, and VGGFace) for image classification when used as data augmentation.
Together with the aforementioned qualitative and quantitative comparisons, the effectiveness and versatility of our method are further identified.

\subsection{Ablation Studies and Parameter Sensitivities}
\label{sec:ablation_studies}
In this part, we ablate different modules to testify the efficacy of each component and investigate the loss weights of $\lambda_{str}$ and $\lambda_{fre}$.
%

\begin{table}
\caption{Ablation studies to probe the efficacy of our proposed techniques.
``full" denotes all proposed modules are used.
}
\vspace{-6pt}
\centering
\resizebox{.8\columnwidth}{!} {
\begin{tabular}{l|rr|rr}
\toprule
\multirow{2}{*}{Method} & \multicolumn{2}{c|}{Flowers} & \multicolumn{2}{c}{Animal Faces} \cr
& FID ($\downarrow$)  & LPIPS ($\uparrow$) &FID ($\downarrow$) & LPIPS ($\uparrow$) \cr 
\cmidrule(r){1-1} \cmidrule(r){2-2} \cmidrule(r){3-3} \cmidrule(r){4-4}  \cmidrule(r){5-5}
LoFGAN + ``full" & \textbf{74.08} & \textbf{0.3983} & \textbf{96.74}  & \textbf{0.5028} \cr
w/o TexMod      & 74.41 & 0.3882 & 97.43  & 0.4970 \cr
w/o StructD     & 78.11 & 0.3952 & 109.43 & 0.5001 \cr
w/o FreD        & 75.80 & 0.3928 & 98.32  & 0.5010 \cr \midrule
WaveGAN + ``full"& \textbf{39.51} & \textbf{0.3970} & \textbf{26.65}  & \textbf{0.5109} \cr
w/o TexMod      & 40.23 & 0.3859 & 26.90  & 0.5069 \cr
w/o StructD     & 41.28 & 0.3956 & 29.82  & 0.5096 \cr
w/o FreD        & 42.04 & 0.3942 & 27.05  & 0.5100 \cr  \bottomrule
\end{tabular}
}
\label{tab:ablation_module}
\end{table}

\noindent \textbf{Module ablation.}
We mute each module and keep other settings unchanged to probe their impacts.
~\cref{tab:ablation_module} presents the qualitative results.
Despite being evaluated on different baselines (\emph{i.e.,} LoFGAN and WaveGAN) and datasets (\emph{i.e.,} Flowers and Animal Faces), the empirical results consistently reflect the efficacy of our proposed techniques.
More precisely, the proposed StrcutD and FreD mainly contribute to the FID score (\emph{e.g.,} from 78.11 or 75.80 to 74.08 on Flowers, respectively), matching our goal to improve overall faithfulness.
By contrast, TexMod pours more attention into improving the synthesis diversity. Namely, removing TexMod leads to severe degradation in the LPIPS score (\emph{e.g.,} from 0.3970 to 0.3859 on Flowers).
Additionally, by combining these techniques, we obtain the best synthesis quality in terms of FID and LPIPS scores.
That is, they complement each other for further improvements.
%


\begin{table}[h]
\centering
\caption{Ablation studies on the loss weights $\lambda_{str}$ and $\lambda_{fre}$.
%
}
\vspace{-6pt}
\centering
\resizebox{.65\linewidth}{!}{
\begin{subtable}
\centering
\begin{tabular}{c c c} \toprule
$\lambda_{str}$ & $\lambda_{fre}$ & FID ($\downarrow$) \\ \midrule
0   & 0 & 4.96 \\ \midrule
0.1 & 0 & 4.89 \\
1   & 0 & \textbf{4.37}  \\
10  & 0 & 5.01 \\
100 & 0 & 49.12 \\
\bottomrule
\end{tabular}
\end{subtable}\hfill
 \begin{subtable}
 \centering
 \hspace{.05in}
\begin{tabular}{c c c } \toprule
$\lambda_{str}$ & $\lambda_{fre}$ & FID ($\downarrow$) \\
\midrule
0   & 1   & 4.72 \\
\midrule
1   & 0.1 & 4.35 \\
1   & 1   & \textbf{4.03} \\
1   & 10  & 4.29 \\
1   & 100 & 8.52 \\
\bottomrule
\end{tabular}
\end{subtable}\hfill
  \label{tab:ablation_loss}
}
\end{table}
\vspace{-5pt}

\noindent \textbf{Constraint strength.}
Recall that StructD and FreD are involved as loss terms for optimization in implementation.
Therefore, here we further perform ablative comparisons on their constraint strength to investigate the parameter sensitivities.
Specifically, we first set $\lambda_{str}$ and $\lambda_{fre}$ to zero to obtain the baseline FID score on the VGGFace dataset.
Then we investigate a proper value for $\lambda_{str}$ in [0.1, 1, 10, 100], wherein $\lambda_{fre}$ is set to 0.
After obtaining an appropriate coefficient for $\lambda_{str}$, we turn to explore $\lambda_{fre}$ in [0.1, 1, 10, 100].
Finally, suitable choices for both $\lambda_{str}$ and $\lambda_{fre}$ could be derived.
Notably, TexMod is not used here to avoid unnecessary impacts.

~\cref{tab:ablation_loss} presents the quantitative results.
We could tell that $\lambda_{str} = \lambda_{str} = 1$ fit best to our goal.
Too small or strong coefficients might either fail to enforce the model to capture corresponding information or surpass other constraints thus leading to imbalanced training.
More results can be found in the appendix.

\subsection{Comparison of Various Numbers of Shots}
In order to investigate the performance of our model under different numbers of input images, we evaluate our model with different numbers of input images, \emph{i.e.,} $K$ $\in$ [3, 5, 7, 9].
We add our techniques on LoFGAN and test on the Flowers dataset here.

\begin{figure}
    \centering
    \includegraphics[width=.40\textwidth]{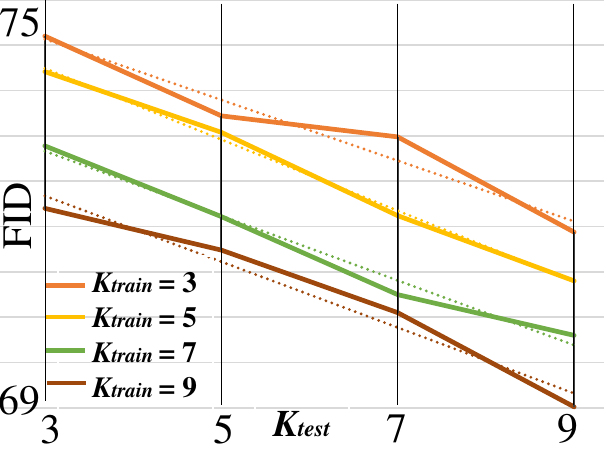}
    \vspace{-6pt}
    \caption{Comparison results under different shots. The dotted lines represent the average slope, demonstrating the overall trend of the FID scores as the sample size increases.}
    \label{fig:various_shots}
\end{figure}

~\cref{fig:various_shots} presents the FID scores under different $K$-shot generation tasks.
We could tell that better synthesis performance could be gained via 1) involving more input images for training, or 2) increasing the number of testing images for evaluation.
Such observation is reasonable as more images provide more semantic variances and meaningful representations for the synthesis.

\subsection{Cross-domain Generation}
Recall that the model is expected to capture the knowledge of learning how to produce novel images instead of mimicking the training distribution.
To further evaluate how well the model could transfer learned knowledge to irrelevant domains, we perform a cross-domain generation here.
Concretely, the model is first trained on the VGGFace dataset.
Then, we input a few images from the Animal Faces dataset for testing.

\begin{figure}
    \centering
    \includegraphics[width=.45\textwidth]{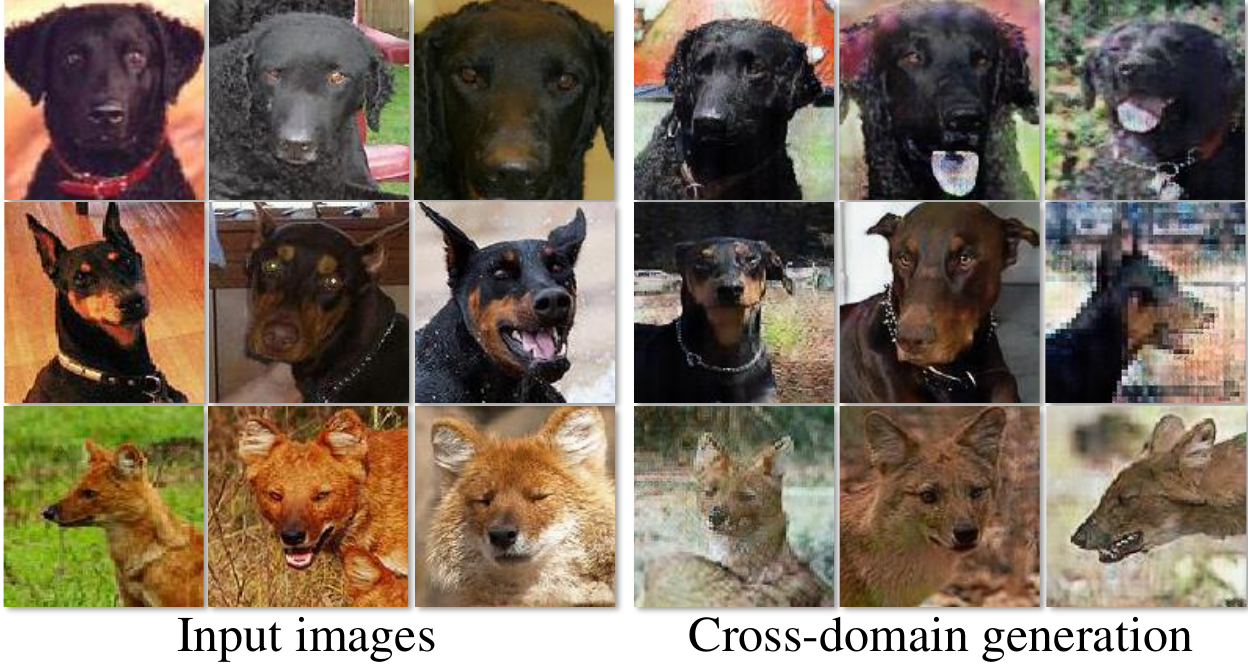}
    \vspace{-6pt}
    \caption{Cross-domain generation results. The model is trained on VGGFace dataset while tested on Animal Faces dataset.}
    \label{fig:cross_domain}
\end{figure}

~\cref{fig:cross_domain} shows the qualitative results.
%
Interestingly, although the synthesis quality slightly drops, our model can still produce acceptable images under such a setting, demonstrating that the model indeed captures the ability of generating rather than memorizing training images.
Quantitative results are provided in appendix.

\section{Conclusion}

In this work, we propose a general few-shot image generation model with two delicate designs, namely textural modulation (TexMod) and structural discrimination (StructD).
Firstly, the representative ability and structural awareness of the discriminator are improved by explicitly providing global guidelines to it, facilitating a more faithful generation.
Secondly, we achieve more fine-grained representation fusion by injecting external semantic layouts into internal textures. 
Additionally, being parameterized by the discriminator's feedback, TexMod is capable of maintaining the synthesis fidelity.
As a result, our model could produce high-quality samples with superior diversity and faithfulness,
and the generated images could be leveraged as augmentation for improving downstream classification tasks.
Furthermore, our proposed techniques complement existing approaches and facilitate cross-domain generation.

\begin{acks}
This work is supported by Shanghai Science and Technology Program under Grant No. 21511100800, Natural Science Foundation of China under Grant No. 62076094, Shanghai Science and Technology Program under Grant No. 20511100600, and Natural Science Foundation of China under Grant No. 62002193.
\end{acks}

\bibliographystyle{ACM-Reference-Format}
\balance
\bibliography{sections/ref}

\newpage
\appendix
\section{Appendix}
In this appendix, we provide more discussion, qualitative, and quantitative results to better illustrate the advancement of our proposed techniques.
Concretely, the limitation and potential future works are discussed, followed by the qualitative comparison between images synthesized by WaveGAN~\cite{wavegan} and our proposed method, and the quantitative results of the cross-domain evaluation.
Additionally, more fine-grained analyses on the parameters $\lambda_{str}$ and $\lambda_{fre}$ are presented.

\noindent \textbf{Limitations.}
Despite achieving substantial improvements on all evaluated datasets, there remain areas for further improvement in our proposed model
Specifically, the model's performance might suffer when generalizing to datasets with significant class variances, such as ImageNet~\cite{ImageNet}.
Moreover, the cross-domain generation capability is still suboptimal, particularly when the domain gap is substantial, like transferring from the human face domain to natural flowers.
Finally, the synthesis quality of our model on extremely limited data amounts, such as one-shot generation tasks, can be further enhanced.
These limitations might be approached in the following two ways:
1) Incorporating various data augmentation techniques (\emph{e.g.,} adaptive data augmentation (ADA) in~\cite{stylegan2-ada} and differentiable augmentation from~\cite{DiffAug}) to enlarge the sample amount of one-shot generation tasks.
2) Exploring additional modules to capture more internal distributional information for the generation tasks. 
Despite these limitations, our model offers promising alternatives to enhance few-shot image generation and downstream classification problems. 

\noindent \textbf{Future works.}
%
In addition to addressing the above limitations, we plan to perform our further studies on few-shot image generation in the following two ways:
First, collecting high-resolution benchmarks for experimental evaluation as the resolutions of existing popular datasets are relatively low (128 $\times$ 128 $\times$ 3).
This would effectively advance the field of few-shot image generation and promote more promising applications in various domains.
Second, investigating the performance of diffusion models~\cite{latentdiffusion} on few-shot image generation since diffusion models have become the new trend of the generative community.
Accordingly, it is crucial to incorporate the excellent attributes of diffusion models (\emph{e.g.,} simple objectives and training stability) into few-shot image generation.

\begin{figure}[t]
    \centering
    \includegraphics[width=\linewidth]{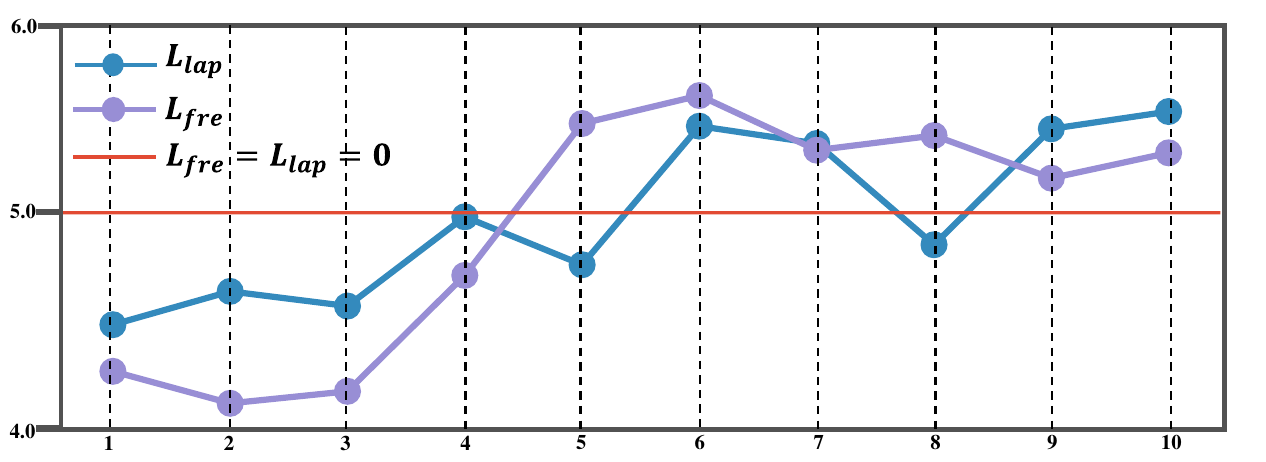}
    \caption{Parameter sensitivity analysis on $\lambda_{str}$ and $\lambda_{fre}$.}
    \label{fig:parameter}
\end{figure}

\noindent \textbf{Parameter sensitivity.}
In order to investigate the upper bound of our proposed model under different values and pairs of $\lambda_{str}$ and $\lambda_{fre}$, we conduct further investigations within the range of [1, 10].
~\cref{fig:parameter} presents the parameter sensitivities of $\lambda_{str}$ and $\lambda_{fre}$.
Similarly, too strong coefficients might make the additional losses surpass other constraints, leading to imbalanced training and performance degradation.
Moreover, better performance can be obtained with different values in the range of [1, 3], demonstrating the effectiveness of our proposed techniques.
Furthermore, the rationality of our setting in the main experiments $\lambda_{str} = \lambda_{fre} =1$ is identified.

\begin{figure*}[t]
    \centering
    \includegraphics[width=\textwidth]{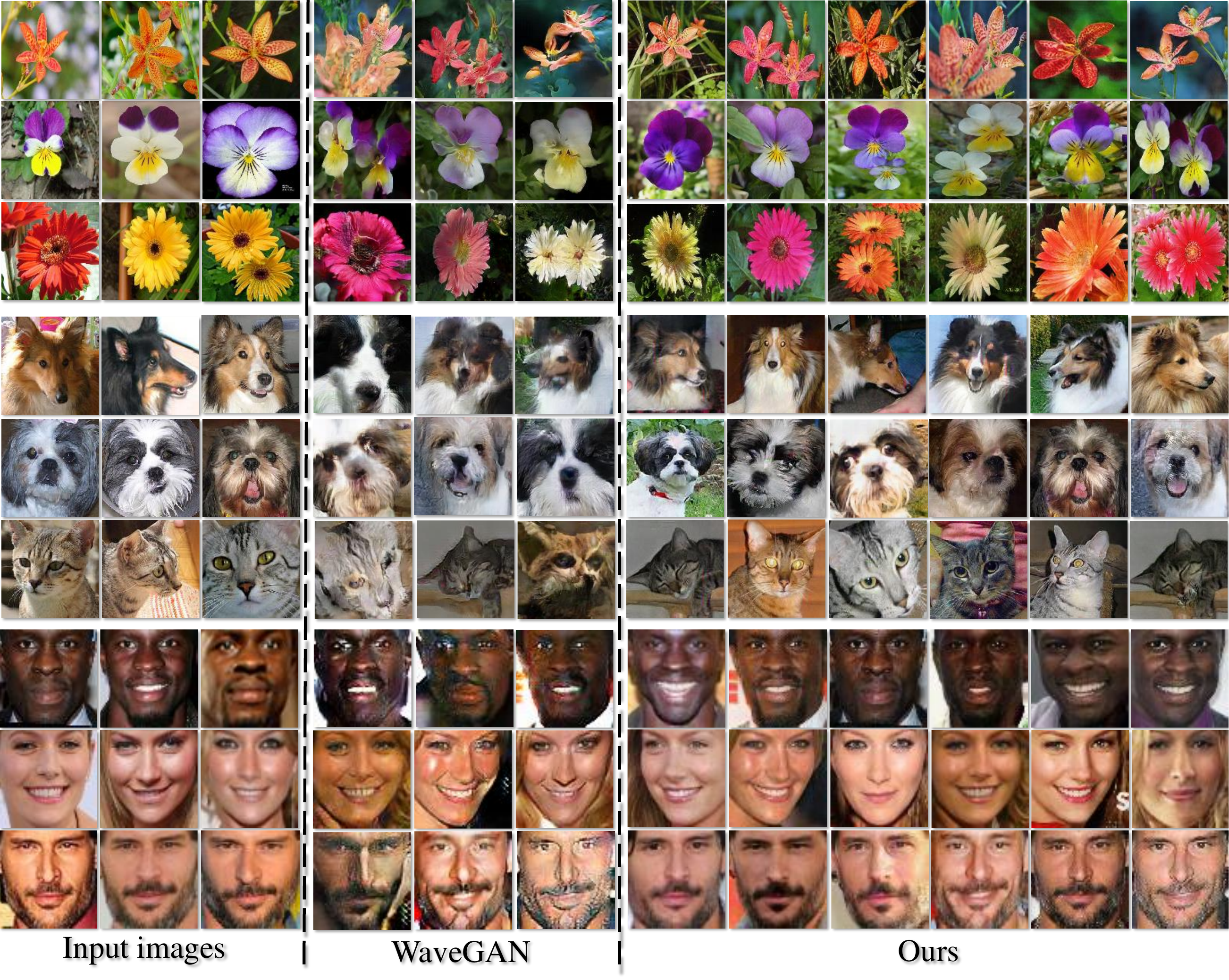}
    \caption{Qualitative comparison results of our method with WaveGAN. Images produced by our model performs better in term of the global structure (\emph{e.g.,} the outline and shape of petals and the coherence of Animal Faces) and semantic variance (\emph{e.g.,} different hair colors of Animal Faces and various expression of Human Face.)}
    \label{fig:qualitative_wavegan}
\end{figure*}

\noindent \textbf{Additional qualitative results.}
~\cref{fig:qualitative_wavegan} presents the qualitative comparison results of our proposed method and WaveGAN~\cite{wavegan}.
Akin to the results in~\cref{fig:qualitative}, these results demonstrate that the images generated by our proposed model exhibit superior fine-grained semantic details, overall structures, and authenticity compared to those generated by WaveGAN.
Such observation further highlights the significant improvements in generation quality achieved by our proposed techniques.
Notably, ~\cref{fig:qualitative} and~\cref{fig:qualitative_wavegan} exhibit some similarities between the results of LoFGAN (\emph{resp.} WaveGAN) and our proposed method due to the use of identical input images during testing.
As a result, the generated output images may exhibit some common features, such as the arrangement of flower petals, the fur color and texture of animal faces, and the facial expressions of human faces.
Nevertheless, it could be seen from these results that the images generated by our proposed approach are characterized by a greater degree of photorealism and visual plausibility.

\noindent \textbf{Additional quantitative results.}
Here we provide the quantitative results of the cross-domain generation experiments with all datasets combinations.
To be more specific, the model is first trained on one domain (\emph{e.g.,} VGGFace) and then tested on another domain (\emph{e.g.,} Animal Faces), while other settings remained consistent with the main experiments.
~\cref{tab:cross_domain_metric} presents the quantitative results.
It is evident that the synthesis performance deteriorates when the training and testing data are from different domains, particularly when the domain gap is substantial (\emph{e.g.,} transferring from Flowers to Animal Faces and VGGFace). 
Nonetheless, our proposed techniques effectively improve the transfer performance under different baselines, further emphasizing the compatibility and flexibility of our method.

\begin{table}[h]
\caption{FID comparison results of cross-domain evaluation.
The model is trained on one source domain dataset (\emph{e.g.,} Flowers) and tested on another target domain dataset (\emph{e.g.,} Animal Faces/VGGFace).
All results are obtained in three-shot settings.
} 
\centering
\resizebox{\columnwidth}{!} {
\begin{tabular}{l|cc|cc|cc}
\toprule
\multirow{2}{*}{Method} & \multicolumn{2}{c|}{Flowers} & \multicolumn{2}{c|}{Animal Faces} & \multicolumn{2}{c}{VGGFace}\cr
                & Animal Faces & VGGFace & Flowers & VGGFace & Flowers & Animal Faces \cr
\cmidrule(r){1-1} \cmidrule(r){2-2} \cmidrule(r){3-3} \cmidrule(r){4-4}  \cmidrule(r){5-5} \cmidrule(r){6-6} \cmidrule(r){7-7}
LoFGAN          & 158.82          & 34.44          & 101.92          & 26.42           & 95.04           & 124.64 \cr
+ Ours          & \textbf{150.09} & \textbf{30.12} & \textbf{99.67}  & \textbf{23.59}  & \textbf{93.46}  & \textbf{119.99} \cr \midrule
WaveGAN         & 56.32           & 16.27          & 89.87           & 12.19           & 68.43           & 59.05 \cr
+ Ours          & \textbf{48.21}  & \textbf{14.35} & \textbf{78.46}  & \textbf{9.61}   & \textbf{65.71}  & \textbf{55.62} \cr \bottomrule
\end{tabular}
}
\label{tab:cross_domain_metric}
\end{table}


\end{document}